\documentclass[runningheads]{llncs}
\usepackage[T1]{fontenc}
\usepackage{graphicx}
\usepackage{amssymb}
\usepackage{hyperref}
\usepackage{dsfont}
\usepackage{epstopdf}
\usepackage{listings}
\usepackage{amsmath}
\usepackage{multirow}
\usepackage{soul}
\usepackage{booktabs}
\usepackage{bbding}
\usepackage{makecell}
\usepackage{stfloats}
\usepackage{makecell}
\usepackage{xspace}
\usepackage{url}
\usepackage{wrapfig}
\usepackage{multirow}
\usepackage{makecell}

\usepackage[misc]{ifsym}

\usepackage{rotating}
\usepackage{graphicx}
\usepackage{amsmath}
\usepackage{booktabs} 
\usepackage{makecell}
\usepackage{multirow}
\usepackage[table]{xcolor}

\usepackage{color}

\institute{}
\begin{document}
\title{MedGround-R1: Advancing Medical Image Grounding via Spatial-Semantic Rewarded \\Group Relative Policy Optimization}
\titlerunning{MedGround-R1}
\authorrunning{H. Xu et al.}

\newcommand*\samethanks[1][\value{footnote}]{\footnotemark[#1]}
\author{Huihui Xu\inst{1,3} \textsuperscript{*}  \and
Yuanpeng Nie\inst{4} \textsuperscript{*}  \and
Hualiang Wang \inst{3} \and
Ying Chen\inst{1}   \and
Wei Li\inst{1}   \and  \\
Junzhi Ning\inst{1}  \and
Lihao Liu\inst{1}  \and
Hongqiu Wang\inst{5}  \and 
Lei Zhu\inst{3, 5} \and
Jiyao Liu\inst{1,6} \and \\
Xiaomeng Li\inst{3} \textsuperscript{$\dagger$} \and
Junjun He\inst{1,2} \textsuperscript{$\dagger$} 
} 
\institute{
Shanghai Artificial Intelligence Laboratory, Shanghai, China \\ \email{hejunjun@pjlab.org.cn}\and Shanghai Innovation Institute, Shanghai, China \and
The Hong Kong University of Science and Technology, Hong Kong SAR, China \and Department of Nephrology, The Seventh Affiliated Hospital, \\Sun Yat-sen University, Shenzhen, China \and The Hong Kong University of Science and Technology (Guangzhou), China \and Fudan University, Shanghai, China
}
\maketitle              % typeset the header of the contribution
\footnotetext[1]{\textsuperscript{*}  Equal Contribution}
\footnotetext[2]{\textsuperscript{$\dagger$} Corresponding Authors}

\begin{abstract}
Medical Image Grounding (MIG), which involves localizing specific regions in medical images based on textual descriptions, requires models to not only perceive regions but also deduce spatial relationships of these regions. Existing Vision-Language Models (VLMs) for MIG often rely on Supervised Fine-Tuning (SFT) with large amounts of Chain-of-Thought (CoT) reasoning annotations, which are expensive and time-consuming to acquire. Recently, DeepSeek-R1 demonstrated that Large Language Models (LLMs) can acquire reasoning abilities through Group Relative Policy Optimization (GRPO) without requiring CoT annotations. In this paper, we adapt the GRPO reinforcement learning framework to VLMs for Medical Image Grounding. We propose the \textbf{Spatial-Semantic Rewarded Group Relative Policy Optimization} to train the model without CoT reasoning annotations. Specifically, we introduce \textbf{Spatial-Semantic Rewards}, which combine spatial accuracy reward and semantic consistency reward to provide nuanced feedback for both spatially positive and negative completions. Additionally, we propose to use the \textbf{Chain-of-Box} template, which integrates visual information of referring bounding boxes into the \texttt{<think>} reasoning process, enabling the model to explicitly reason about spatial regions during intermediate steps. Experiments on three datasets MS-CXR, ChestX-ray8, and M3D-RefSeg demonstrate that our method achieves state-of-the-art performance in Medical Image Grounding. Ablation studies further validate the effectiveness of each component in our approach. Code, checkpoints, and datasets are available at \url{https://github.com/bio-mlhui/MedGround-R1}

\keywords{ Medical Image Grounding \and Spatial-Semantic Rewarded Group Relative Policy Optimization \and DeepSeek-R1 \and Referring Expression Comprehension}
\end{abstract}

\section{Introduction}
Medical image grounding~\cite{lvit,mscxr,wang2024interpretable} is a critical task in computer-aided diagnosis~\cite{lgrnet,hu2024ophclip}, requiring models to localize specific regions in medical images based on textual descriptions~\cite{shadow}. Unlike general image grounding, this task in the medical domain often requires complex reasoning about anatomical structures, pathological features, and their relationships, making it essential for models to not only localize accurately but also reason effectively and meaningfully. 
For instance, grounding a phrase like ``enlarged lymph node near the left lung'' requires understanding spatial relationships and medical context, which goes beyond simple pattern recognition~\cite{tian2023deep,rainmamba} in medical images.
This reasoning capability is crucial for accurate and interpretable predictions, especially in complex medical scenarios where regions of interest may overlap or have ambiguous boundaries.
However, existing methods~\cite{MedRPG,lvit,causalclipseg} often struggle to incorporate such reasoning capabilities, limiting their performance in complex medical tasks.

Existing Vision-Language Models (VLMs) for MIG~\cite{referground} primarily rely on supervised finetuning (SFT) with large amounts of CoT reasoning annotations, which are expensive and time-consuming to acquire. Moreover, these methods typically focus on end-to-end localization without explicitly modelling the reasoning process, leading to suboptimal performance in complex scenarios. For example, they may fail to distinguish between semantically similar pathological regions or misinterpret spatial relationships, especially when dealing with ambiguous or overlapping structures. While recent works~\cite{MedRG} attempt to integrate reasoning through multi-stage pipelines, they often require extensive labelled data which are expensive to acquire.

Recently, DeepSeek-R1~\cite{deepseek} and DeepSeek-Math~\cite{shao2024deepseekmath} introduced Group Relative Policy Optimization (GRPO), a reinforcement learning framework that enables models to develop reasoning capabilities without the need for Chain-of-Thought (CoT) annotations, which achieves comparable and even better performance than OpenAI o1 series. By leveraging a self-evolution strategy through RL, GRPO allows models to autonomously discover reasoning patterns, such as self-verification and reflection, while significantly reducing the dependency on supervised data. This breakthrough opens up new possibilities for image grounding in computer-aided diagnosis~\cite{serp,chen2025slidechat,yu2021metal}, where reasoning is crucial, but labeled data is scarce~\cite{xu2024online}.

To this end, in this paper, 1) we propose the first RL-based framework for Medical Image Grounding, adapting GRPO technique to train VLMs without CoT annotations. 2) We introduce a combination of \textbf{Spatial-Semantic Rewards} to provide nuanced relative evaluations for spatially negative completions. 3) We propose to use the novel \textbf{Chain-of-Box} reasoning template, which explicitly integrates visual information of referring bounding boxes into the \texttt{<think>} process, allowing VLMs to reason with both visual and textual context during intermediate steps. 4) We conduct comprehensive experiments on three publicly available medical datasets, including MS-CXR~\cite{MedRPG}, ChestX-ray8~\cite{ray8}, and M3D-RefSeg~
\cite{m3d}, demonstrating that our method achieves state-of-the-art performance on MIG both quantitatively and qualitatively. Ablation studies further validate the effectiveness of each component.

\section{Method}
\begin{figure}[t]
\caption{\textbf{A Training Step of the Proposed Spatial-Semantic Group Relative Policy Optimization Framework for MIG.}
 $*$ stands for the model is frozen during training.}
\begin{center}
\includegraphics[width=1.0\linewidth]{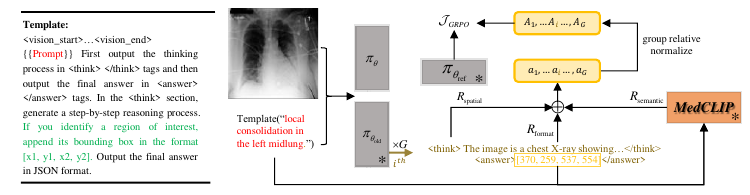}
\end{center}
\label{fig:framework}
\end{figure}
As shown in Fig.\ref{fig:framework}, at each training step, given a medical image and a referring expression, an old policy model $\pi_{\text{old}}$, saved $k$ steps before, is used to randomly sample a group of $G$ different completions. For each completion, we compute three types of rewards. The Format Reward $R_{\text{format}}$ uses regular expression to evaluate whether the completion adheres to the format predefined in the template. The Spatial Reward $R_{\text{spatial}}$ calculates the Intersection over Union (IoU) between the predicted bounding box and the ground truth box, which can be seen as the spatial accuracy. The Semantic Reward $R_{\text{semantic}}$ leverages a frozen MedCLIP~\cite{wang2022medclip,li2025closer,hu2024ophclip} to assess the semantic similarity of the bounding box region of interest (ROI) and the referring expression for each completion. The sum of these three rewards are then normalized to compute each of their \textit{relative advantage} $A_i$. To preserve the original pretrained knowledge, a frozen reference model $\pi_{\text{ref}}$ is devised. The optimization of the current model is performed using the Group Relative Policy Optimization (GRPO) objective $\mathcal{J}_{\text{GRPO}}$~\cite{deepseek,shao2024deepseekmath}. The green part of the template corresponds to the Chain-of-Box prompt. 

It should be noted that unlike previous VLM-based methods~\cite{referground}, our framework does not require any CoT annotation or reasoning data during training. The ground truth reference box is only required.

\subsection{Preliminaries of Group Relative Policy Optimization.}
DeepSeek-R1~\cite{deepseek} is the first open-sourced LLM that achieves comparable or even superior performance to closed models like OpenAI's o1 series. 
The core advancement behind DeepSeek-R1 is Group Relative Policy Optimization (GRPO)~\cite{shao2024deepseekmath,deepseek}, a reinforcement learning algorithm designed to enhance the reasoning capabilities of LLMs without the requirement of large amounts of CoT annotations.

Unlike approaches that rely on supervised fine-tuning (SFT) or extensive Chain-of-Thought (CoT) annotations, GRPO leverages the inherent potential of LLMs to self-evolve through a pure reinforcement learning process. By directly applying RL to the base model, GRPO enables the model to autonomously develop sophisticated reasoning behaviors, such as self-verification and reflection. 

Formally, given a problem $q$, GRPO first samples a set of $G$ different completions $\{s_i\}_{i=1}^{G}$ from an old model $\pi_{\theta_\text{old}}$ saved $k$ steps before. A reward model $R$, usually another LLM or any scalar function, is devised to evaluate the \textit{relative advantage} of each solution:
 \begin{equation}
\begin{split}
    r_i = \mathcal{R}(s_i, q), \quad A_i = \frac{r_i - \text{mean}(\{r_1, r_2, \ldots, r_G\})}{\text{std}(\{r_1, r_2, \ldots, r_G\})},
\end{split}
\end{equation}

By denoting $\pi_\theta(s_i|q)$ as the posterior of $s_i$ predicted by $\pi_\theta$, the following objective is
considered to be maximized:
\begin{align}
    \mathcal{J}_{GRPO}(\theta; q) &= \mathbb{E}_{\{s_i\}_{i=1}^G \sim \pi_{\theta_{old}}(S|q)} \frac{1}{G}\sum_{i=1}^G A_i \frac{\pi_\theta(s_i |q)}{\pi_{\theta_{old}}(s_i |q)},
\end{align}
where if a completion $s_i$ has stronger relative advantage $A_i$, $\pi_\theta$ is expected to be more probable to generate $s_i$ than $\pi_{\theta_{\text{old}}}$. To preserve the original pretrained knowledge, a reference model $\pi_{\theta_{\text{ref}}}$, usually the model before RL, is utilized:
\begin{align}
    \mathcal{J}_{GRPO}(\theta; q) &= \mathbb{E}_{\{s_i\}_{i=1}^G \sim \pi_{\theta_{old}}(S|q)}
    \frac{1}{G}\sum_{i=1}^G A_i \frac{\pi_\theta(s_i |q)}{\pi_{\theta_{old}}(s_i |q)} - \beta \mathbb{D}_{KL}\{\pi_{\theta} (s_i|q) || \pi_{ref} (s_i|q)\},
\end{align}
where $\beta$ is a hyper-parameter for KL-divergence. To stabilize training, clipping~\cite{schulman2017proximal,shao2024deepseekmath} is introduced to formulate the final objective:
% \begin{equation}
% \begin{split}
%     &\mathcal{J}_{GRPO}(\theta) = \mathbb{E}_{\{s_i\}_{i=1}^G \sim \pi_{\theta_{old}}(S|q)}  \\
%     & \frac{1}{G}\sum_{i=1}^G  A_i\left( \min \left( \frac{\pi_\theta(s_i |q)}{\pi_{\theta_{old}}(s_i |q)}, \text{clip} \left( \frac{\pi_\theta(s_i |q)}{\pi_{\theta_{old}}(s_i |q)}, 1 - \epsilon, 1 + \epsilon \right)  \right) - \beta \mathbb{D}_{KL}\{\pi_{\theta} (s_i|q) || \pi_{ref} (s_i|q)\},
% \end{split}
% \label{eq:GRPO-obj}
% \end{equation}
\begin{equation}
\begin{aligned}
\mathcal{J}_{GRPO}(\theta) &= \mathbb{E}_{\{s_i\}_{i=1}^G \sim \pi_{\theta_{old}}(S|q)} \\
&\quad \frac{1}{G}\sum_{i=1}^G A_i \left( \min \left( \frac{\pi_\theta(s_i |q)}{\pi_{\theta_{old}}(s_i |q)}, \text{clip} \left( \frac{\pi_\theta(s_i |q)}{\pi_{\theta_{old}}(s_i |q)}, 1 - \epsilon, 1 + \epsilon \right) \right) \right. \\
&\quad \quad \left. - \beta \mathbb{D}_{KL}\{\pi_{\theta} (s_i|q) || \pi_{ref} (s_i|q)\} \right)
\end{aligned}
\label{eq:GRPO-obj}
\end{equation}
\noindent where $\epsilon$ is the clipping hyperparameter.
The formulation in Eq.~\ref{eq:GRPO-obj} enables GRPO to efficiently optimize the policy model by leveraging group-based relative rewards, eliminating the need for a separate critic model~\cite{schulman2017proximal} and also significantly reducing computational overhead.

% format, iou, clip, 
\subsection{Format Reward for Medical Image Grounding.}
\noindent\textbf{Format Reward.} Following DeepSeek-R1~\cite{deepseek}, the format reward \( R_{\text{format}} \) ensures the model's completion adheres to a predefined structure. Specifically, the completion must enclose its reasoning process within \texttt{<think>} and \texttt{</think>} tags, followed by the final answer within \texttt{<answer>} and \texttt{</answer>} tags. Moreover, in the grounding task, the \texttt{<answer>} tag must additionally contain a bounding box in the format \texttt{[$x_1$, $y_1$, $x_2$, $y_2$]}. Both rewards are implemented by regular expressions.\footnote{ {\tiny
\texttt{r"<think>.*?</think>\textbackslash s*<answer>.*?\textbackslash\{.*\textbackslash[\textbackslash d+,\textbackslash s*\textbackslash d+,\textbackslash s*\textbackslash d+,\textbackslash s*\textbackslash d+\textbackslash].*\textbackslash\}.*?</answer>"}}} \footnote{ {\tiny
\texttt{r"\textbackslash[(\textbackslash d+),\textbackslash s*(\textbackslash d+),\textbackslash s*(\textbackslash d+),\textbackslash s*(\textbackslash d+)\textbackslash]"}}} If a completion matches both patterns, indicating that it adheres to the required format and contains a valid bounding box, the reward is 1. Otherwise, the reward is 0.

\subsection{Spatial-Semantic Consistency Rewards}

\noindent\textbf{IoU as Spatial Consistency Reward.} To evaluate the spatial accuracy of the predicted bounding box for each completion, Intersection over Union (IoU) \( R_{\text{IoU}} \) between the ground truth box is calculated. If the IoU exceeds a threshold of 0.5, the reward is 1; otherwise, the reward is 0. This reward $R_{\text{spatial}}$ can be seen as measuring the spatial consistency of each completion with respect to the ground truth.

Although $R_{\text{spatial}}$ is sufficient to evaluate the \textit{spatial} accuracy, it can not evaluate their \textit{semantic} consistency with the ground truth. Our insight is that completions that are spatially-negative maybe semantically-positive, in the sense that some of them may belong to the same semantic class with the ground truth and should therefore be assigned larger advantage than those neither spatially nor semantically positive. 

To address this limitation, we introduce the \textbf{Semantic Consistency Reward} $R_{\text{semantic}}$. We leverage a frozen MedCLIP~\cite{wang2022medclip}, to compute the cosine semantic similarity between the cropped bounding box ROI feature of the completion and the referring expression feature. The $R_{\text{semantic}}$ reward provides a continuous reward signal that captures the semantic alignment between the predicted region and the referring text, even when the predicted box does not spatially overlap with the ground truth. 

\subsection{Chain-of-Box Template for Medical Image Grounding}
In contrast to DeepSeek-R1~\cite{deepseek} where the reasoning process within \texttt{<think>} tags is purely textual information, we propose the \textbf{Chain-of-Box} prompt designed for VLMs in medical image grounding task. Our key insight is that the medical image grounding task requires the model to reason about spatial local regions and their relationships in the image during its thinking steps. The reasoning process essentially involves shifting attention to different regions of interest (ROIs).

As illustrated in Fig.~\ref{fig:framework}, whenever the model references a ROI region in the image, it explicitly appends the corresponding bounding box coordinates \texttt{[$x_1$, $y_1$, $x_2$, $y_2$]} after the region text. This \textbf{Chain-of-Box} approach ensures the visual information is seamlessly integrated into the reasoning context, enabling VLMs to perform multimodal reasoning effectively.

\section{Experiments}
 
\noindent\textbf{Datasets.} We evaluate our method on three publicly available datasets. The first dataset is \textbf{MS-CXR}~\cite{mscxr}, derived from MIMIC-CXR~\cite{MIMIC-CXR}, which contains 1,153 image-phrase-box triples. Following MedRPG~\cite{MedRPG}, we preprocess the data to ensure that each phrase query corresponds to a single bounding box, resulting in 890 samples. The second dataset is \textbf{ChestX-ray8}~\cite{ray8}, a large-scale dataset for diagnosing eight common chest diseases. It includes 984 pathology images with hand-labeled bounding boxes. We adopt the preprocessing approach from MedRPG~\cite{MedRPG}, using category labels as phrase queries to construct image-text-box triplets. The third dataset is \textbf{M3D-RefSeg}~\cite{m3d}, which consists of 2,778 mask-text-volume triplets extracted from the Totalsegmentator~\cite{totalseg} dataset. To adapt for 2D grounding tasks, we extract the frontal slice from each 3D volume and convert the segmentation mask into the bounding box. 

\noindent\textbf{Compared Methods.} We compare our method against seven recent state-of-the-art approaches, including \textbf{LViT}~\cite{lvit}, \textbf{MedRPG}~\cite{MedRPG}, \textbf{CausalCLIPSeg}~\cite{causalclipseg}, \textbf{ChEX}~\cite{ChEX}, \textbf{GuideDecoder}~\cite{guidedecoder}, \textbf{RecLMIS}~\cite{RecLMIS}, and one VLM-based method \textbf{BiRD}~\cite{referground}. For segmentation-based methods~\cite{lvit,causalclipseg,guidedecoder,RecLMIS}, we convert their output masks into bounding boxes to allow comparison. Each method adopts the hyperparameter settings from their original papers. Notably, BiRD~\cite{referground} is fine-tuned based on Qwen2-VL in its original paper. To ensure a fair comparison, we retrain BiRD based on Qwen2.5-VL~\cite{qwen25vl} same as our setting.

\noindent\textbf{Evaluation Metrics.}  We follow the protocol established in \textbf{MedRPG}~\cite{MedRPG} to evaluate all methods. Specifically, we report \textbf{Accuracy (Acc)}, where a predicted region is considered correct if its Intersection over Union (IoU) with the ground-truth bounding box exceeds 0.5. Additionally, we report the \textbf{mean IoU (mIoU)} metric to provide a more comprehensive comparison of localization precision. 

\noindent\textbf{Implementation Details.} Our model is based on finetuning Qwen2.5-VL~\cite{qwen25vl} under the proposed reinforcement learning framework. We use a batch size of 1 per GPU with gradient accumulation over 2 steps. The model is trained for 5K steps with an initial learning rate of \(1 \times 10^{-6}\). We employ mixed precision training with bfloat16, gradient checkpointing, and Flash Attention to optimize computation efficiency. We set $G=4$, $\beta=0.04$ in our experiments. The maximum completion length is set to 256. More ablation studies on hyperparameters are demonstrated in Sec.~\ref{sec:ablation}.
\begin{table*}[!t]
  \centering
  \caption{\textbf{Quantitative comparisons of all methods on three datasets.}}
  \label{tab:sota_compare}
  \begin{tabular}{cc|cc|cc|cc} 
        \toprule \toprule
        \multirow{2}{*}{Method} & \multirow{2}{*}{Publication} & \multicolumn{2}{c}{MS-CXR}  & \multicolumn{2}{c}{ChestX-ray8}  & \multicolumn{2}{c}{M3D-RefSeg} \\
        \cmidrule(lr){3-8} & & mIoU$\uparrow$ & Acc$\uparrow$ & mIoU$\uparrow$ & Acc$\uparrow$ & mIoU$\uparrow$ & Acc$\uparrow$ \\
        \hline
        Lvit~\cite{lvit}  & TMI'23                      &  52.67  &   66.94 &  33.50  & 35.05 & 40.22 & 48.79 \\
        GuideDecoder~\cite{guidedecoder} & MICCAI'23    &  56.21  &   67.04 &  32.70  & 34.68 & 42.19 & 51.68 \\
        MedRPG~\cite{MedRPG}  &  MICCAI'23              &  59.37  &   69.86 &  34.59  & 38.02 & 41.25 & 50.35 \\
        RecLMIS~\cite{RecLMIS} & TMI'24                 &  62.04  &   72.49 &  33.05  & 40.46 & 46.71 & 57.22\\
        ChEX~\cite{ChEX} & ECCV'24                      &  61.23  &   70.15 &  38.49  & 41.60 & 45.83 & 56.01 \\
    CausalCLIPSeg~\cite{causalclipseg} &  MICCAI'24     &  62.77  &   73.18 &  40.12  & 46.02 & 47.11 & 58.36 \\
        \hline
        BiRD~\cite{referground}   & MICCAI'24           &  73.33  &   76.05 &  48.22  & 50.12 & 52.02 & 70.18 \\
        \hline
        \textbf{Ours}  & \textbf{-- --}                          & \textbf{79.02}  & \textbf{83.12}  & \textbf{53.12} & \textbf{62.18} & \textbf{60.10} & \textbf{74.66} \\
        \bottomrule \bottomrule
  \end{tabular}
\end{table*}

\subsection{Comparisons with SOTA methods}
\begin{figure}[t]
\centering
\caption{\textbf{Qualitative comparisons on MS-CXR. }The blue dashed boxes are ground truth bounding boxes.}
\label{fig:quality}
\includegraphics[width=0.95\linewidth]{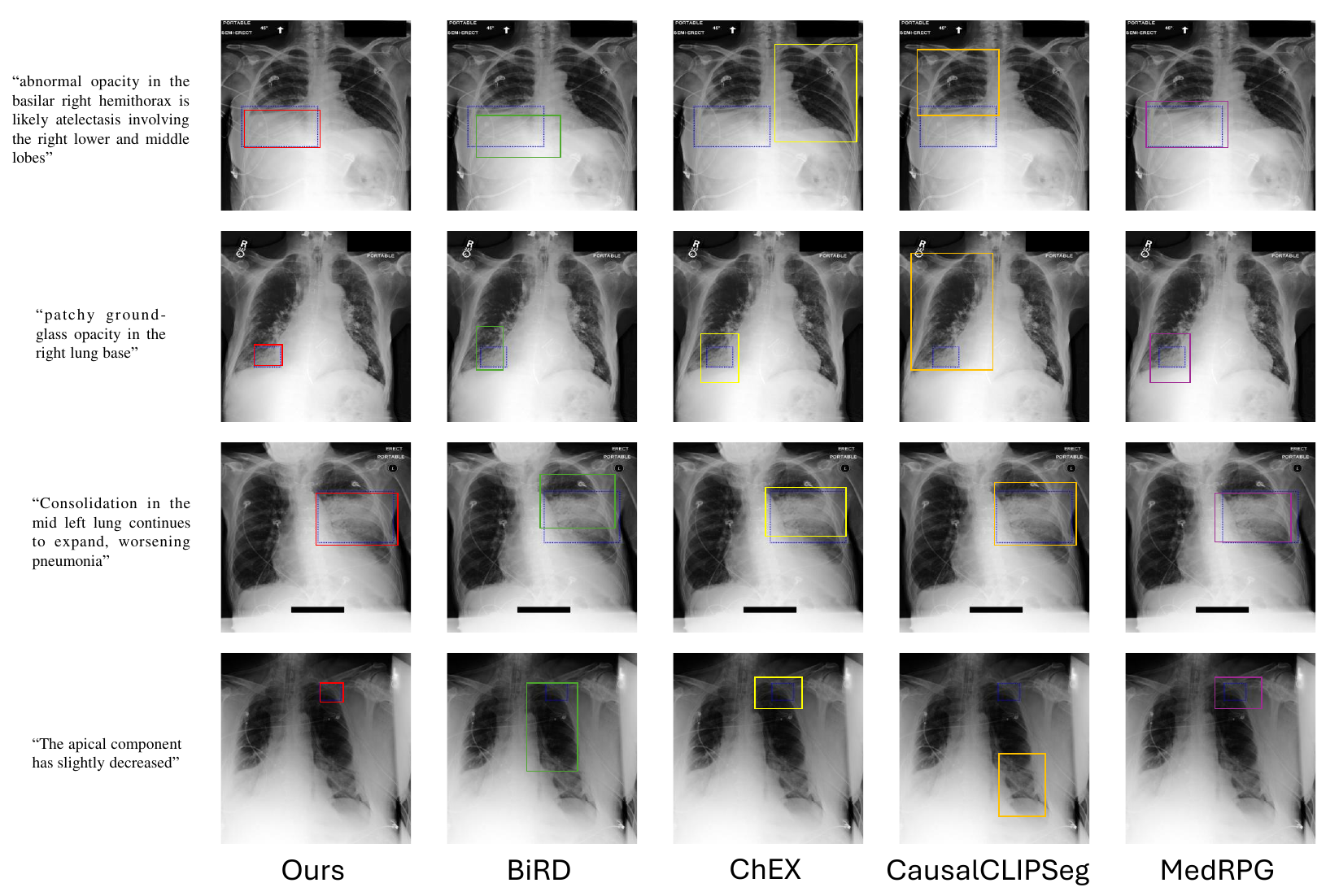}
\end{figure}

Our method achieves state-of-the-art performance across all three datasets, demonstrating significant improvements over existing approaches. On MS-CXR, we achieve a mean Intersection over Union (mIoU) of \textbf{79.02} and an accuracy (Acc) of \textbf{83.12}, outperforming the previous best method, BiRD, by 5.69 and 7.07 points, respectively. On ChestX-ray8, our method attains an mIoU of \textbf{53.12} and an accuracy of \textbf{62.18}, surpassing BiRD by 4.90 and 12.06 points. For the M3D-RefSeg dataset, our method achieves an mIoU of \textbf{60.10} and an accuracy of \textbf{74.66}, outperforming BiRD by 8.08 and 4.48 points. The consistent performance gains across all datasets validate the effectiveness of our proposed reinforcement learning framework, particularly the integration of Spatial-Semantic Rewards and the Chain-of-Box reasoning template, which together enhance the grounding accuracy. 

\noindent\textbf{Visual Comparison}  
As shown in Fig.~\ref{fig:quality}, qualitative results further demonstrate the superiority of our method. Visual comparisons reveal that the bounding boxes predicted by our model are significantly more aligned with the ground truth (GT) compared to other methods. This alignment is also evident in cases with complex anatomical structures or ambiguous referring expressions, such as row 1 and 3, where our model's ability to reason spatially and semantically ensures precise localization. For example, in cases where multiple regions of interest overlap, our method consistently identifies the correct region, while other methods often produce inaccurate or overly broad bounding boxes.
\vspace{-0.4cm}
\begin{table}[!h] 
\centering
\begin{minipage}[!h]{0.3\linewidth}
    \caption{Number of Sampled Completions.} \label{tab:ab_g}
    \centering
    \resizebox{0.5\textwidth}{!}{%
    \begin{tabular}{c|cc}
    \toprule \toprule
    $G$ & mIoU$\uparrow$ & Acc$\uparrow$\\
    \midrule
    2 & 77.40 & 81.05 \\
    4 & 79.02 & 83.12 \\
    6 & 80.16 & 83.43 \\
    8 & 80.77 & 84.01  \\
    \bottomrule \bottomrule
    \end{tabular}%
    }
\end{minipage}\hfill
\begin{minipage}[!h]{0.31\linewidth}
    \caption{Reward Modeling.} \label{tab:ab_reward}
    \centering
    \resizebox{\textwidth}{!}{%
    \begin{tabular}{ccc|cc}
     \toprule \toprule
    $R_{\text{format}}$ & $R_{\text{spatial}}$ & $R_{\text{semantic}}$ & mIoU$\uparrow$ & Acc$\uparrow$\\
    \midrule
    \checkmark & \checkmark & \checkmark        & 79.02 & 83.12\\
    $\times$          & \checkmark & \checkmark & 61.22 & 68.74\\
    \checkmark & $\times$        & \checkmark   & 2.66 & 4.30\\
    \checkmark & \checkmark & $\times$          & 72.14 & 76.46\\
    \bottomrule \bottomrule
    \end{tabular}%
    }
\end{minipage}\hfill
\begin{minipage}[!h]{0.35\linewidth}
\caption{Finetuning Regime} \label{tab:ab_regime}
\centering
\resizebox{0.8\textwidth}{!}{
\begin{tabular}{c|cc}
\toprule \toprule
Regime  & mIoU$\uparrow$ & IoU$\uparrow$ \\
\midrule
SFT-1k                     & 68.65 & 74.18 \\
SFT-3k                     & 56.26  & 67.50 \\
GRPO-1k                      & 64.25   & 73.86      \\
GRPO-3k                      & 70.58  & 75.61  \\
GRPO-5k                      & 77.58  & 81.09      \\
\midrule
\textbf{GRPO-5K w. Chain-of-Box}       & 79.02 & 83.12  \\
\bottomrule \bottomrule
\end{tabular}%
}
\end{minipage}
\vspace{-0.3cm}
\end{table}

\subsection{Ablation Study}
\label{sec:ablation}
\noindent\textbf{Different Setting of $G$.} $G$ stands for the number of sampled completions at each training step. All models are finetuned with 5K steps on MS-CXR. As shown in Tab.~\ref{tab:ab_g}, the performance improves as $G$ increases. However, larger values of $G$ also lead to higher computational costs at each training step, including the time required for sampling and computing rewards for each completion. This trade-off between performance and computational efficiency must be carefully considered..

\noindent\textbf{Different Rewards Combination.} $R_{\text{format}}$, $R_{\text{spatial}}$, and $R_{\text{semantic}}$ play different role in policy decision. All models are finetuned with 5K steps on MS-CXR. As shown in Tab.~\ref{tab:ab_reward}, removing the format reward $R_{\text{format}}$ causes performance drop. We found some outputs are invalid and do not adhere to the standard box pattern. Without spatial reward, which directly evaluates the accuracy between the prediction and the ground truth box, performance collapses significantly. Removing the semantic reward \( R_{\text{semantic}} \) results in a moderate drop, as it provides nuanced relative evaluations for spatially-negative completions, distinguishing semantically related predictions from unrelated ones. These results confirm that all three rewards are essential and effective for robust performance.

\noindent\textbf{SFT, GRPO, Chain-of-Box Template.} As shown in Tab.~\ref{tab:ab_regime}, we finetune the VLM under two regimes: SFT (maximize the next-token probability of the ground-truth box) and GRPO with varying training steps on MS-CXR. Due to the small training set, SFT tends to overfit, as evidenced by the performance drops from 1k to 3k steps. In contrast, GRPO maintains stable performance. Moreover, compared with the original GRPO template used in DeepSeek-R1-Zero, the Chain-of-box template is shown to be effective in improving performance for multimodal reasoning in MIG.

\section{Conclusion}
In this work, we present a reinforcement learning (RL) framework for Medical Image Grounding, leveraging Group Relative Policy Optimization (GRPO). To adapt DeepSeek-R1 GRPO to VLMs, we introduce two key innovations: (1) a combination of \textbf{Spatial-Semantic Rewards} to provide nuanced relative advantage for the spatially-negative completions, and (2) a \textbf{Chain-of-Box} prompt template that explicitly integrates visual information into the intermediate reasoning steps. Extensive experiments on three datasets demonstrate the superiority of our method, achieving state-of-the-art performance. Ablation studies further validate the effectiveness of each component.

\begin{credits}
\subsubsection{\ackname} 
This work was supported by the National Key R\&D Program of China (2022ZD0160101, 2022ZD0160102), the National Natural Science Foundation of China (Grant No.62272450), Shanghai Artificial Intelligence Laboratory, a research grant from the Joint Research Scheme (JRS) under the National Natural Science Foundation of China (NSFC), the Research Grants Council (RGC) of Hong Kong (Project No. N\_HKUST654/24), as well as a grant from the RGC of the Hong Kong Special Administrative Region, China (Project No. R6005-24).
\subsubsection{\discintname}
The authors have no competing interests to declare that are
relevant to the content of this article. 

\end{credits}

\newpage

\bibliographystyle{splncs04}
\bibliography{Paper-1222}
\end{document}